\pdfoutput=1

\documentclass[11pt]{article}

\usepackage{emnlp2021}
\usepackage{amsmath}
\usepackage{amssymb}
\usepackage{xspace}

\usepackage{times}
\usepackage{latexsym}
\usepackage{graphicx}
\usepackage{booktabs}
\usepackage{amsthm}

\newtheorem{definition}{Definition}
\newtheorem{example}{Example}
\newtheorem{proposition}{Proposition}

\newcommand{\vinfo}{$\mathcal{V}$-information\xspace}
\newcommand{\xuvinfo}{Xu-$\mathcal{V}$-information\xspace}
\newcommand{\dtrain}{\mathcal{D}_{\text{tr}}\xspace}
\newcommand{\dtest}{\mathcal{D}_{\text{te}}\xspace}

\usepackage[T1]{fontenc}

\usepackage[utf8]{inputenc}

\usepackage{microtype}

\title{Conditional probing: measuring usable information beyond a baseline}

\newcommand{\AnD}{\hskip 1.5em plus 1fil minus 0.5em}
\author{John Hewitt \AnD Kawin Ethayarajh \AnD Percy Liang \AnD Christopher D. Manning  \\
Department of Computer Science \\
Stanford University \\
  \texttt{\{johnhew,kawin,pliang,manning\}@cs.stanford.edu}}

\begin{document}
\maketitle
\begin{abstract}
\microtypesetup{enable}
Probing experiments investigate the extent to which neural representations make properties---like part-of-speech---predictable. 
One suggests that a representation encodes a property if probing that representation produces higher accuracy than probing a baseline representation like non-contextual word embeddings.
Instead of using baselines as a point of comparison, we're interested in measuring information that is contained in the representation but not in the baseline. %
For example, current methods can detect when a representation is more useful than the word identity (a baseline) for predicting part-of-speech; however, they cannot detect when the representation is predictive of just the aspects of part-of-speech \textit{not} explainable by the word identity.
In this work, we extend a theory of \textit{usable} information called $\mathcal{V}$-information and propose \textit{conditional probing}, which explicitly conditions on the information in the baseline.
In a case study, we find that after conditioning on non-contextual word embeddings, properties like part-of-speech are accessible at deeper layers of a network than previously thought.
\end{abstract}

\section{Introduction}
Neural language models have become the foundation for modern NLP systems \citep{devlin2019bidirectional,radford2018improving}, but what they understand about language, and how they represent that knowledge, is still poorly understood \citep{belinkov2019analysis,rogers2020primer}.
The \textit{probing} methodology grapples with these questions by relating neural representations to well-understood properties.
Probing analyzes a representation by using it as input into a supervised classifier, which is trained to predict a property, such as part-of-speech \citep{shi2016does,ettinger2016probing,alain2016understanding,adi2017finegrained,belinkov2021probing}. %

One suggests that a representation encodes a property of interest if probing that representation produces higher accuracy than probing a baseline representation like non-contextual word embeddings.
However, consider a representation that encodes \textit{only} the part-of-speech tags that aren't determined by the word identity.
Probing would report that this representation encodes \textit{less about part-of-speech} than the non-contextual word baseline, since ambiguity is relatively rare.
Yet, this representation clearly encodes interesting aspects of part-of-speech.
How can we capture this?

In this work, we present a simple probing method to explicitly condition on a baseline.\footnote{Our code is available at \url{https://github.com/john-hewitt/conditional-probing}.}
For a representation and a baseline, our method trains two probes: (1) on just the baseline, and (2) on the concatenation of the baseline and the representation. %
The performance of probe (1) is then subtracted from that of probe (2).
We call this process \textit{conditional probing}.
Intuitively, the representation is not penalized for \textit{lacking} aspects of the property accessible in the baseline.

We then theoretically ground our probing methodology in \vinfo, a theory of \textit{usable} information introduced by \citet{xu2020theory} that we additionally extend to multiple predictive variables.
We use \vinfo instead of mutual information \cite{shannon1948mathematical,pimentel2020information} because any injective deterministic transformation of the input has the same mutual information as the input.
For example, a representation that maps each unique sentence to a unique integer must have the same mutual information with any property as does BERT's representation of that sentence, yet the latter is more useful.
In contrast, \vinfo is defined with respect to a family of functions $\mathcal{V}$ that map one random variable to (a probability distribution over) another. %
\vinfo can be constructed by deterministic transformations that make a property more accessible to the functions in the family.
We show that conditional probing provides an estimate of \textit{conditional \vinfo} $I_{\mathcal{V}}(\text{repr} \rightarrow \text{property } | \text{ baseline})$.

In a case study, we answer an open question posed by \citet{hewitt2019control}: how are the aspects of linguistic properties that \textit{aren't explainable by the input layer} accessible across the rest of the layers of the network? %
We find that the part-of-speech information not attributable to the input layer remains accessible much deeper into the layers of ELMo \cite{peters2018deep} and RoBERTa \cite{liu2019roberta} than the overall property, a fact previously obscured by the gradual loss across layers of the aspects attributable to the input layer.
For the other properties, conditioning on the input layer does not change the trends across layers.

\section{Conditional \vinfo Probing}
In this section, we describe probing methods and introduce conditional probing.
We then review \vinfo and use it to ground probing.

\subsection{Probing setup}
We start with some notation.
Let $X\in\mathcal{X}$ be a random variable taking the value of a sequence of tokens.
Let $\phi(X)$ be a representation resulting from a deterministic function of $X$; for example, the representation of a single token from the sequence in a layer of BERT \cite{devlin2019bidirectional}.
Let $Y\in\mathcal{Y}$ be a property (e.g., part-of-speech of a particular token), and $\mathcal{V}$ a \textit{probe family}, that is, a set of functions $\{f_\theta: \theta \in \mathbb{R}^{p}\}$, where $f_\theta : z \rightarrow \mathcal{P}(\mathcal{Y})$ maps inputs $z$ to probability distributions over the space of the label.\footnote{We discuss mild constraints on the form that $\mathcal{V}$ can take in the Appendix. Common probe families including linear models and feed-forward networks meet the constraints.}
The input $z\in\mathbb{R}^m$ may be in the space of $\phi(X)$, that is, $\mathbb{R}^d$, or another space, e.g., if the probe takes the concatenation of two representations.
In each experiment, a training dataset $\mathcal{D}_\text{tr} = \{(x_i,y_i)\}_i$ is used to estimate $\theta$, and the probe and representation are evaluated on a separate dataset $\mathcal{D}_{\text{te}} = \{(x_i,y_i)\}_i$.
We refer to the result of this evaluation on some representation $R$ as $\text{Perf}(R)$.

\subsection{Baselined probing}
Let $B \in \mathbb{R}^d$ be a random variable representing a baseline (e.g., non-contextual word embedding of a particular token.)
A common strategy in probing is to take the difference between a probe performance on the representation and on the baseline \cite{zhang2018language}; we call this \textbf{baselined probing performance}:
\begin{align}
\text{Perf}(\phi(X)) - \text{Perf}(B).
\end{align}
This difference in performances estimates how much more \textit{accessible} $Y$ is in $\phi(X)$ than in the baseline $B$, under probe family $\mathcal{V}$.

But what if $B$ and $\phi(X)$ capture distinct aspects of $Y$?
For example, consider if $\phi(X)$ captures parts-of-speech that aren't the most common label for a given word identity, while $B$ captures parts-of-speech that \text{are} the most common for the word identity.
Baselined probing will indicate that $\phi(X)$ explains less about $Y$ than the baseline, a ``negative'' probing result.
But clearly $\phi(X)$ captures an interesting aspect of $Y$; we aim to design a method that measures just what $\phi(X)$ \textit{contributes beyond $B$} in predicting $Y$, not what $B$ has and $\phi(X)$ lacks.

\subsection{Our proposal: conditional probing}
In our proposed method, we again train two probes; each is the concatenation of two representations of size $d$, so we let $z\in\mathbb{R}^{2d}$.
The first probe takes as input $[B; \phi(X)]$, that is, the concatenation of $B$ to the representation $\phi(X)$ that we're studying.
The second probe takes as input $[B; \mathbf{0}]$, that is, the concatenation of $B$ to the $\mathbf{0}$ vector.
The conditional probing method takes the difference of the two probe performances, which we call \textbf{conditional probing performance}:
\begin{align}
\text{Perf}([B;\phi(X)]) - \text{Perf}([B;\mathbf{0}]).
\end{align}
Including $B$ in the probe with $\phi(X)$ means that $\phi(X)$ only needs to contribute what is missing from $B$.
In the second probe, the $\mathbf{0}$ is used as a placeholder, representing the lack of knowledge of $\phi(X)$; its performance is subtracted so that $\phi(X)$ isn't given credit for what's explainable by $B$.\footnote{The value $\mathbf{0}$ is arbitrary; any constant can be used, or one can train the probe on just $B$.} %

\subsection{\vinfo}
$\mathcal{V}$-information is a theory of \textit{usable} information---that is, how much knowledge of random variable $Y$ can be extracted from r.v. $R$ when using functions in $\mathcal{V}$, called a \textit{predictive family} \cite{xu2020theory}.
Intuitively, by explicitly considering computational constraints, $\mathcal{V}$-information can be \emph{constructed} by computation, in particular when said computation makes a variable easier to predict.
If $\mathcal{V}$ is the set of all functions from the space of $R$ to the set of probability distributions over the space of $Y$, then \vinfo is mutual information \cite{xu2020theory}.
However, if the predictive family is the set of all functions, then no representation is more useful than another provided they are related by a bijection.
By specifying a $\mathcal{V}$, one makes a hypothesis about the functional form of the relationship between the random variables $R$ and $Y$.
One could let $\mathcal{V}$ be, for example, the set of log-linear models.

Using this predictive family $\mathcal{V}$, one can define the uncertainty we have in $Y$ after observing $R$ as the $\mathcal{V}$-entropy:
\begin{align} \label{eqn_v_entropy}
    H_{\mathcal{V}}(Y|R) = \inf_{f\in\mathcal{V}}\mathbb{E}\big[-\log f[r](y)\big], 
\end{align}
where $f[r]$ produces a probability distribution over the labels. 
Information terms like $I_{\mathcal{V}}(R \rightarrow Y)$ are defined analogous to Shannon information, that is, $I_{\mathcal{V}}(R \rightarrow Y) = H_{\mathcal{V}}(Y) - H_{\mathcal{V}}(Y|R)$.
For brevity, we leave a full formal description, as well as our redefinition of \vinfo to multiple predictive variables, to the appendix.

\subsection{Probing estimates $\mathcal{V}$-information}
With a particular performance metric, baselined probing estimates a difference of \vinfo quantities.
Intuitively, probing specifies a function family $\mathcal{V}$, training data is used to find $f\in V$ that best predicts $Y$ from $\phi(X)$ (the infimum in Equation~\ref{eqn_defn_ventropy}), and we then evaluate how well $Y$ is predicted.
If we use the negative cross-entropy loss as the $\text{Perf}$ function, then \textbf{baselined probing} estimates
\begin{align*}
I_{\mathcal{V}}(\phi(X)\rightarrow Y) - I_{\mathcal{V}}(B\rightarrow Y),
\end{align*}
the difference of two $\mathcal{V}$-information quantities.
This theory provides methodological best practices as well: the form of the family $\mathcal{V}$ should be chosen for theory-external reasons,\footnote{There are also PAC bounds \cite{valiant1984theory} on the estimation error for \vinfo \cite{xu2020theory}; simpler families $\mathcal{V}$ with lower Rademacher complexity result in better bounds.}
 and since the probe training process is approximating the infimum in Equation~\ref{eqn_v_entropy}, we're not concerned with sample efficiency. %

Baselined probing appears in existing information-theoretic probing work: \citet{pimentel2020information} define conditional mutual information quantities wherein a lossy transformation $c(\cdot)$ is performed on the sentence (like choosing a single word), and an estimate of the gain from knowing the rest of the sentence is provided; $I(\phi(X);Y|c(\phi(X)))= I(X;Y|c(X))$.\footnote{Equality depends on the injectivity of $\phi$; otherwise knowing the representation $\phi(X)$ may be strictly less informative than knowing $X$.}
Methodologically, despite being a conditional information, this is identical to baselined probing, training one probe on just $\phi(X)$ and another on just $c(\phi(X))$.\footnote{
This is because of the data processing inequality and the fact that $c(\phi(X))$ is a deterministic function of $\phi(X)$.
}

\subsection{Estimating conditional information}
Inspired by the transparent connections between $\mathcal{V}$-information and probes, we ask what the $\mathcal{V}$-information analogue of conditioning on a variable in a mutual information, that is, $I(X,Y|B)$.
To do this, we extend %
$\mathcal{V}$-information to multiple predictive variables, and design conditional probing (as presented) to estimate
\begin{align*}
I_{\mathcal{V}}(\phi(X)& \rightarrow Y | B) \\&= H_{\mathcal{V}}(Y|B) - H_{\mathcal{V}}(Y|B,\phi(X)),
\end{align*}
thus having the interpretation of probing what $\phi(X)$ explains about $Y$ apart from what's already explained by $B$ (as can be accessed by functions in $\mathcal{V}$).
Methodologically, the innovation is in providing $B$ to the probe on $\phi(X)$, so that the information accessible in $B$ need not be accessible in $\phi(X)$.

\section{Related Work}
Probing---mechanically simple, but philosophically hard to interpret \cite{belinkov2021probing}---has led to a number of information-theoretic interpretations. %

\citet{pimentel2020information} claimed that probing should be seen as estimating mutual information $I(\phi(X);Y)$ between representations and labels.
This raises an issue, which \citet{pimentel2020information} notes: due to the data processing inequality, the MI between the representation of a sentence (from e.g., BERT) and a label is upper-bounded by the MI between the sentence itself and the label.
Both an encrypted document $X$ and an unencrypted version $\phi(X)$ provide the same mutual information with the topic of the document $Y$.
This is because MI allows unbounded work in using $X$ to predict $Y$, including the enormous amount of work (likely) required to decrypt it without the secret key.
Intuitively, we understand that $\phi(X)$ is more useful than $X$, and that this is because the function $\phi$ performs useful ``work''  for us.
Likewise, BERT can perform useful work to make interesting properties more accessible. 
While \citet{pimentel2020information} conclude from the data processing inequality that probing is not meaningful, we conclude that estimating mutual information is not the goal of probing.

\citet{voita2020informationtheoretic} propose a new probing-like methodology, \textit{minimum description length (MDL) probing}, to measure the number of bits required to transmit both the specification of the probe and the specification of labels. Intuitively, a representation that allows for more efficient communication of labels (and probes used to help perform that communication) has done useful ``work'' for us.
\citet{voita2020informationtheoretic} found that by using their methods, probing practitioners could pay less attention to the exact functional form of the probe.
\vinfo and MDL probing complement each other; \vinfo does not measure sample efficiency of learning a mapping from $\phi(X)$ to $Y$, instead focusing solely on how well any function from a specific family (like linear models) allows one to predict $Y$ from $\phi(X)$.
Further, in practice, one must choose a family to optimize over even in MDL probing; the complexity penalty of communicating the member of the family is analogous to choosing $\mathcal{V}$.
Further, our contribution of conditional probing is orthogonal to the choice of probing methodology; it could be used with MDL probing as well.

\vinfo places the functional form of the probe front-and-center as a \textit{hypothesis} about how structure is encoded.
This intuition is already popular in probing.
For example, \citet{hewitt2019structural} proposed that syntax trees may emerge as squared Euclidean distance under a linear transformation.
Further work refined this, showing that a better structural hypothesis may be hyperbolic \cite{chen2021probing} axis-aligned after scaling \cite{limisiewicz-marecek-2021-introducing}, or an attention-inspired kernel space \cite{white-etal-2021-non}.

In this work, we intentionally avoid claims as to the ``correct'' functional family $\mathcal{V}$ to be used in conditional probing.
Some work has argued for simple probe families \cite{hewitt2019control,alain2016understanding}, others for complex families  \cite{pimentel2020information,hou2021birds}. %
\citet{pimentel2020pareto} argues for choosing multiple points along an axis of expressivity, while \citet{cao2021low} define the family through the weights of the neural network.
Other work performs structural analysis of representations without direct supervision \cite{saphra2018understanding,wu2020perturbed}.

\citet{hewitt2019control} suggested that differences in ease of identifying the word identity across layers could impede comparisons between the layers; our conditional probing provides a direct solution to this issue by conditioning on the word identity.
\citet{kuncoro2018lstms} and \citet{shapiro2021multilabel} use control tasks, and \citet{rosa2020measuring} measures word-level memorization in probes. 
Finally, under the possible goals of probing proposed by \citet{ivanova2021probing}, we see \vinfo as most useful in \textit{discovering emergent structure}, that is, parsimonious and surprisingly simple relationships between neural representations and complex properties.\looseness=-1

\section{Experiments}
In our experiments, we aim for a case study in understanding how conditioning on the non-contextual embeddings changes trends in the accessibility of linguistic properties across the layers of deep networks.

\subsection{Tasks, models, and data}
\paragraph{Tasks.} We train probes to predict five linguistic properties, roughly arranged in order from lower-level, more concrete properties to higher-level, more abstract properties.
We predict five linguistic properties $Y$: (i) \textbf{upos}: coarse-grained (17-tag) part-of-speech tags \cite{nivre2020universal}, (ii) \textbf{xpos}: fine-grained English-specific part-of-speech tags, (iii) \textbf{dep rel}: the label on the Universal Dependencies edge that governs the word, (iv) \textbf{ner}: named entities, and (v) \textbf{sst2}: sentiment.

\paragraph{Data.}
All of our datasets are composed of English text.
For all tasks except sentiment, we use the Ontonotes v5 corpus \citep{weischedel2013ontonotes}, recreating the splits used in the CoNLL 2012 shared task, as verified against the split statistics provided by \citet{strubell2017fast}.\footnote{In order to provide word vectors for each token in the corpus, we heuristically align the subword tokenizations of RoBERTa with the corpus-specified tokens through character-level alignments, following \citet{tenney2018what}.}\footnote{Ontonotes uses the destructive Penn Treebank tokenization (like replacing brackets \texttt{\{} with \texttt{-LCB-} \cite{marcus1993building}). We perform a heuristic de-tokenization process before subword tokenization to recover some naturalness of the text.}
Since Ontonotes is annotated with constituency parses, not Universal Dependencies, we use the converter provided in CoreNLP \citep{schuster2016enhanced,manning2014stanford}.
For the sentiment annotation, we use the binary GLUE version \cite{wang2018glue} of the the Stanford Sentiment Treebank corpus \citep{socher2013recursive}.
All results are reported on the development sets.

\paragraph{Models.}
We evaluate the popular RoBERTa model \cite{liu2019roberta}, as provided by the HuggingFace Transformers package \citep{wolf2019huggingfaces}, as well as the ELMo model \citep{peters2018deep}, as provided by the AllenNLP package \citep{gardner2017allennlp}.
When multiple RoBERTa subwords are aligned to a single corpus token, we average the subword vector representations.

\paragraph{Probe families.}
For all of our experiments, we choose $\mathcal{V}$ to be the set of affine functions followed by softmax.\footnote{We used the Adam optimizer \cite{kingma2014adam} with starting learning rate 0.001, and multiply the learning rate by $0.5$ after each epoch wherein a new lowest validation loss is not achieved.} For word-level tasks, we have
\begin{align}
f_\theta(\phi_i(X)_j) = \text{softmax}(W\phi_i(X)_j+b)
\end{align}
where $i$ indexes the layer in the network and $j$ indexes the word in the sentence.
For the sentence-level sentiment task, we average over the word-level representations, as 
\begin{align}
f_\theta(\phi_i(X)) =\text{softmax}(W\ \text{avg}(\phi_i(X))+b)
\end{align}

\subsection{Results}

\begin{figure}
  \includegraphics[width=0.49\linewidth]{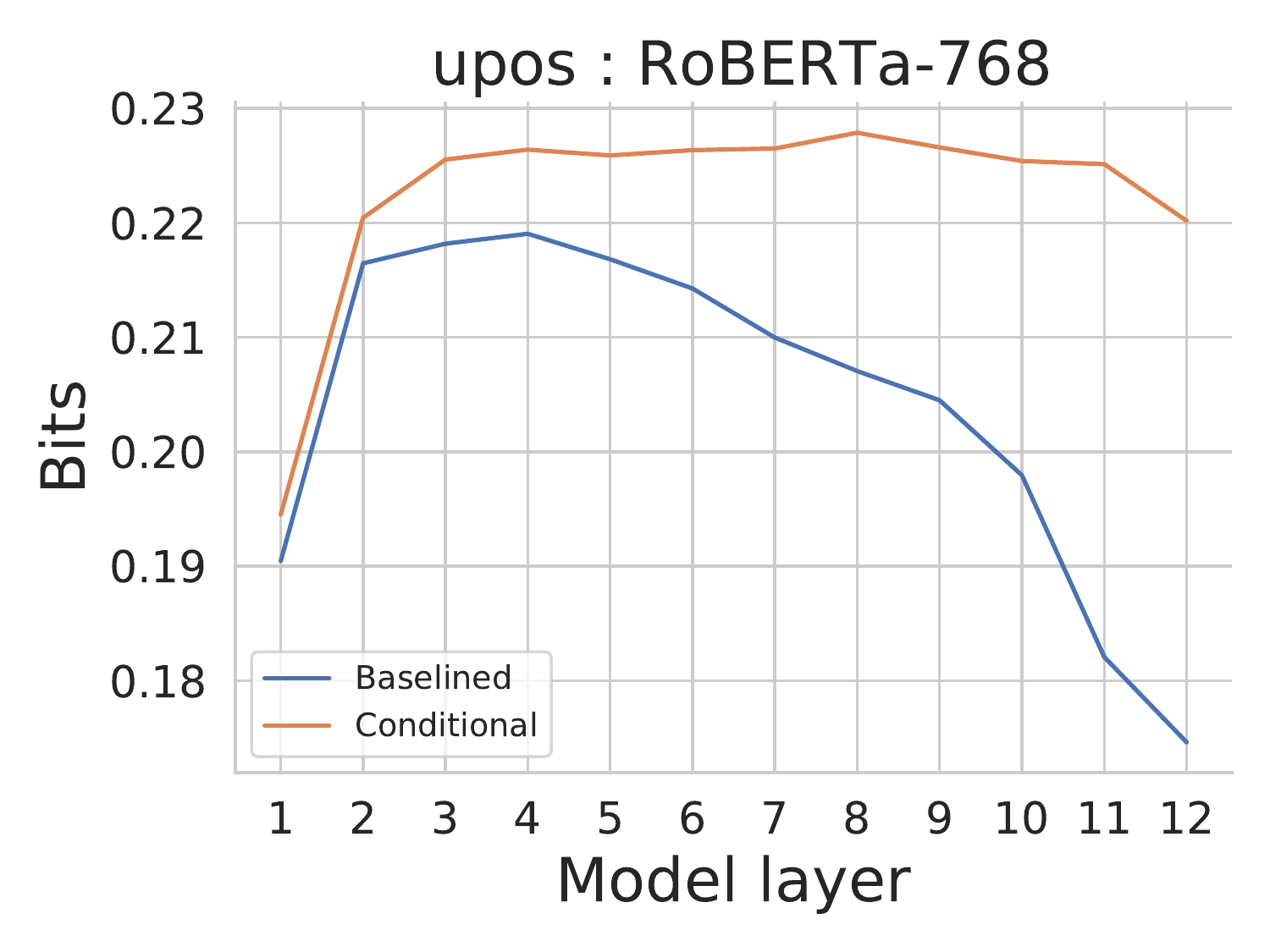}
  \includegraphics[width=0.49\linewidth]{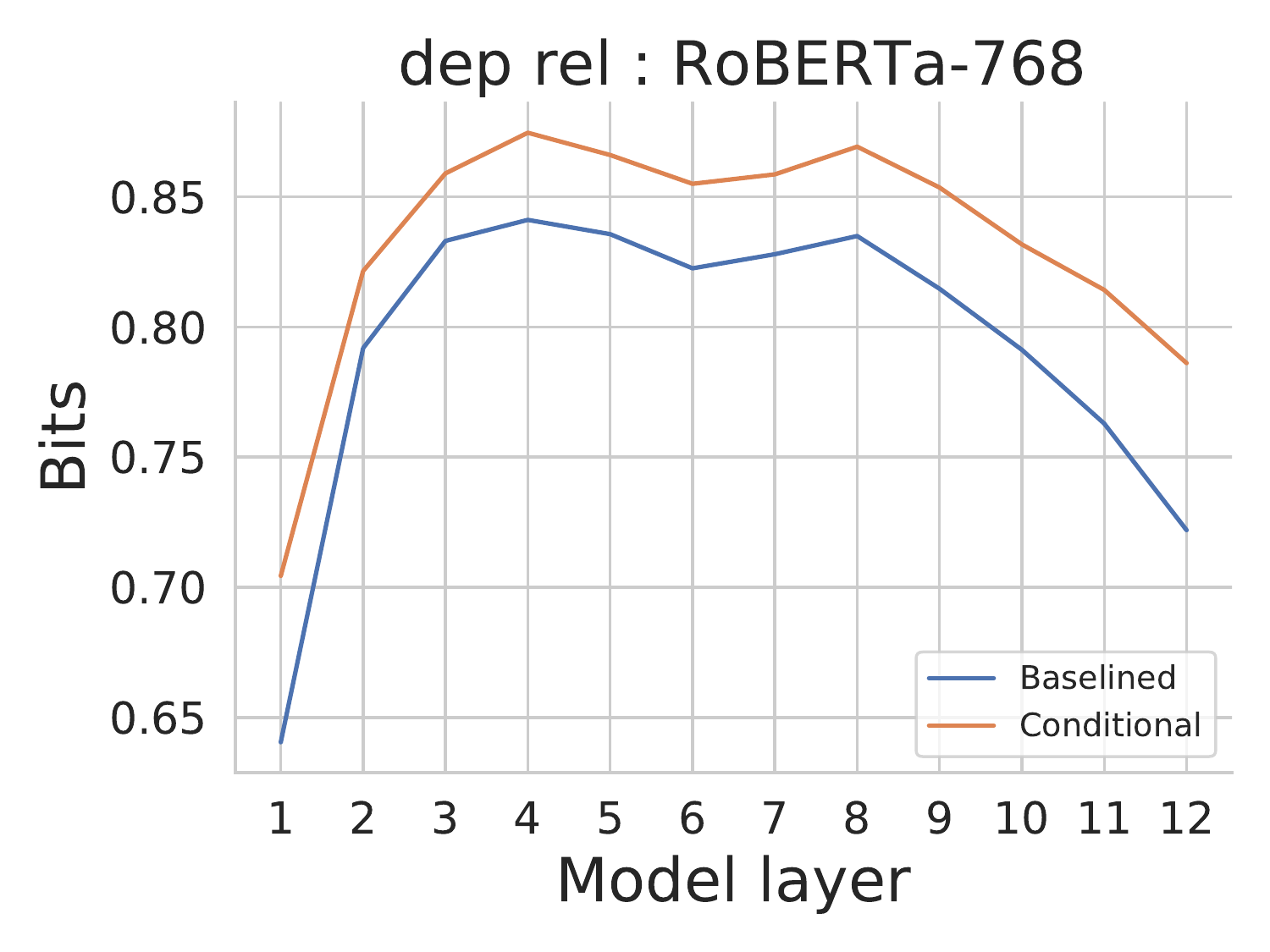}
  \includegraphics[width=0.49\linewidth]{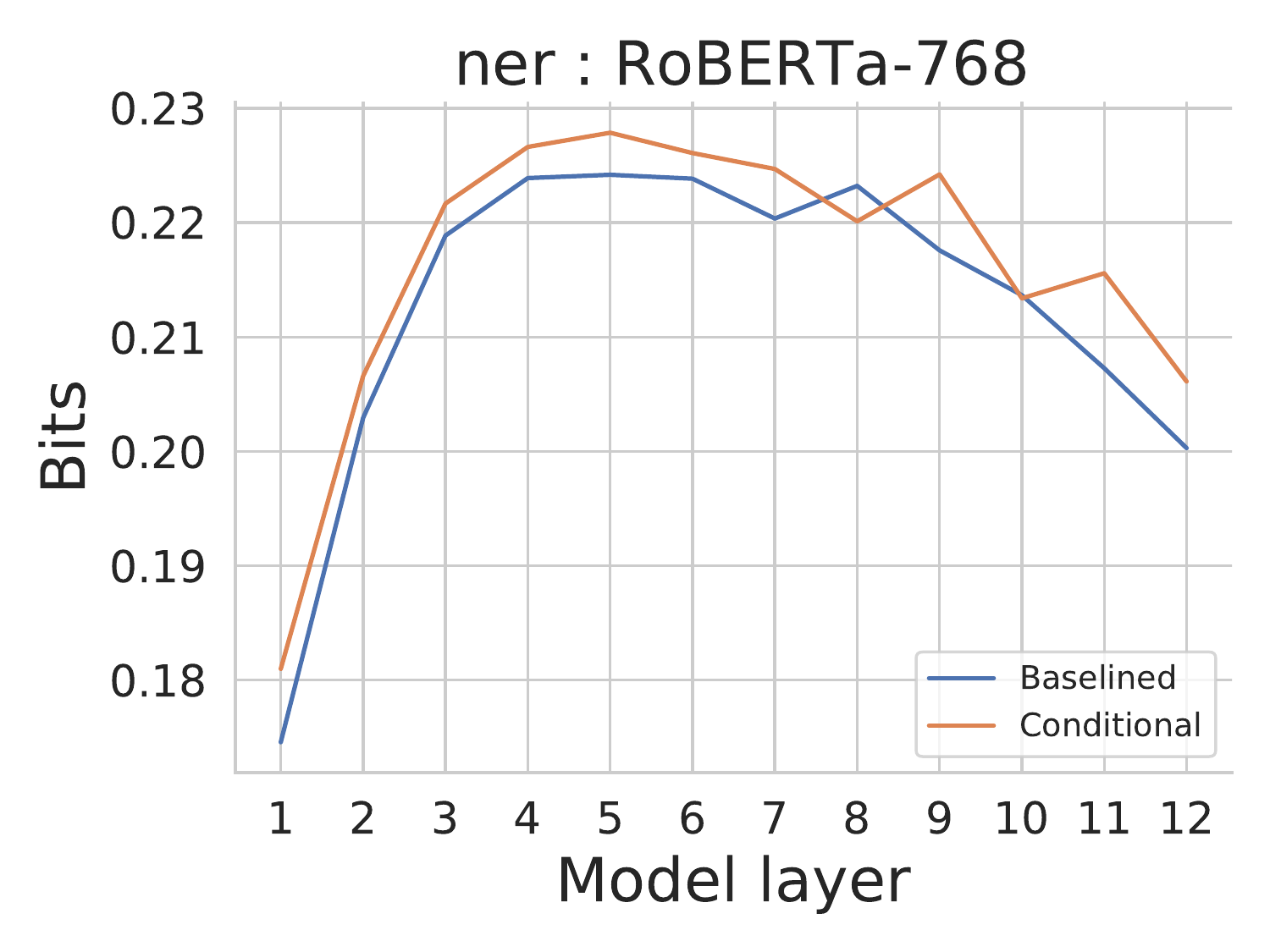}
  \includegraphics[width=0.49\linewidth]{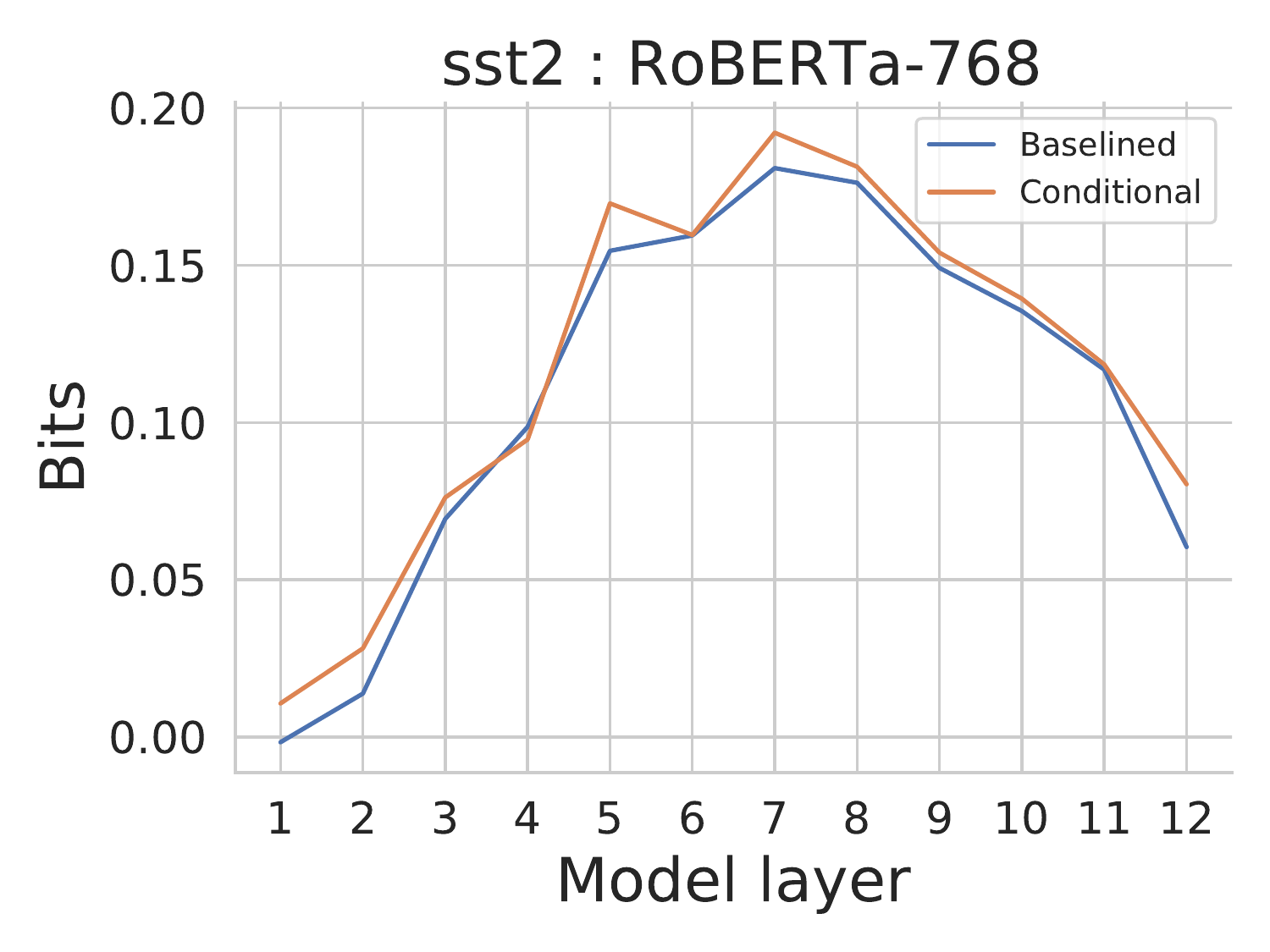}
  \caption{\label{fig_results_bert_roberta_eci_eri}Probing results on RoBERTa. Results are reported in bits of $\mathcal{V}$-information; higher is better.
   }
\end{figure}

\begin{table}
    \centering
    \small
  \begin{tabular}{l c c c c c}

\toprule
& \multicolumn{2}{c}{Baselined} & & \multicolumn{2}{c}{Conditional}\\
\midrule
& $\phi_1$ & $\phi_2$ & & $\phi_1$ & $\phi_2$\\
\cmidrule{2-3}
\cmidrule{5-6}
upos & 0.20  & 0.16 & & 0.22 & 0.20\\
xpos & 0.20  & 0.16 & & 0.21 & 0.20\\
dep rel & 0.99  & 0.81 & & 1.00 & 0.87\\
ner & 0.24  & 0.23 & & 0.25 & 0.24\\
sst2 & 0.18  & 0.13 & & 0.17 & 0.13\\
\bottomrule
     \end{tabular}
     \caption{\label{table_elmo_results}Results on ELMo, reported in bits of $\mathcal{V}$-information; higher is better. $\phi_i$ refers to layer $i$.}
\end{table}

\paragraph{Results on ELMo.}
ELMo has a non-contextual embedding layer $\phi_0$, and two contextual layers $\phi_1$ and $\phi_2$, the output of each of two bidirectional LSTMs \citep{hochreiter1997long}.
Previous work has found that $\phi_1$ contains more syntactic information than $\phi_2$ \cite{peters2018dissecting,zhang2018language}.
Baselined probing performance, in Table~\ref{table_elmo_results}, replicates this finding.
But \citet{hewitt2019control} conjecture that this may be due to accessibility of information from $\phi_0$.
Conditional probing answers shows that when only measuring information not available in $\phi_0$, there is still more syntactic information in $\phi_1$ than $\phi_2$, but the difference is much smaller.

\paragraph{Results on RoBERTa.}
RoBERTa-base is a pretrained Transformer consisting of a word-level embedding layer $\phi_0$ and twelve contextual layers $\phi_i$, each the output of a Transformer encoder block \citep{vaswani2017attention}.
We compare baselined probing performance to conditional probing performance for each layer.
In Figure~\ref{fig_results_bert_roberta_eci_eri}, baselined probing indicates that part-of-speech information decays in later layers.
However, conditional probing shows that information \textit{not} available in $\phi_0$ is maintained into deeper layers in RoBERTa, and only the information already available in $\phi_0$ decays.
In contrast for dependency labels, we find that the difference between layers is lessened after conditioning on $\phi_0$, and for NER and sentiment, conditioning on $\phi_0$ does not change the results.

\section{Conclusion}
In this work, we proposed \textit{conditional probing},  a simple method for conditioning on baselines in probing studies, and grounded the method theoretically in \vinfo.
In a case study, we found that after conditioning on the input layer, usable part-of-speech information remains much deeper into the layers of ELMo and RoBERTa than previously thought, answering an open question from \citet{hewitt2019control}.
Conditional probing is a tool that practitioners can easily use to gain additional insight into representations.\footnote{An executable version of the experiments in this paper is on CodaLab, at this link: \url{https://worksheets.codalab.org/worksheets/0x46190ef741004a43a2676a3b46ea0c76}.}

\bibliography{anthology,custom}
\bibliographystyle{acl_natbib}

\appendix

\section{Multivariable \vinfo} \label{appendix_vinfo}
In this section we introduce Multivariable \vinfo.
\vinfo as introduced by \citet{xu2020theory} was defined in terms of a single predictive variable $X$, and is unwieldy to extend to multiple variables due to its use of a “null” input outside the sample space of $X$ (Section~\ref{appendix_section_xu_ours_relationship}).\footnote{In particular, the null input encodes \textit{not} knowing the value of $X$; technical conditions in the definition of \vinfo as to the behavior of this null value increase in number exponentially with the number of predictive variables.}
Our multivariable V-information removes the use of null variables and naturally captures the multivariable case. 
Consider an agent attempting to predict $Y \in \mathcal{Y}$ from some information sources $X_1,\dots, X_n$, where $X_i \in \mathcal{X}_i$.
Let $\mathcal{P}(\mathcal{Y})$ be the set of all probability distributions over $Y$.

At a given time, the agent may only have access to a subset of the information sources.
Let the known set $C\in\mathcal{C}$ and unknown set $\bar{C}\in\bar{\mathcal{C}}$ be a binary partition of $X_1,\dots,X_n$.
Though the agent isn't given the true value of $\bar{C}$ when predicting $Y$, it is instead provided with a constant value $\bar{a}\in \bar{\mathcal{C}}$, which does not vary with $Y$.\footnote{The exact value of $\bar{a}$ will not matter, as a result of Definition~\ref{defn_predictive_family}.}

We first specify constraints on the set of functions that the agent has at its disposal for predicting Y from X:

\begin{definition}[Multivariable Predictive Family] \label{defn_predictive_family}
Let $\Omega = \{ f : \mathcal{X}_1\times\cdots\times\mathcal{X}_n \rightarrow \mathcal{P}(\mathcal{Y})\}$.
We say that $\mathcal{V} \subseteq \Omega$ is a predictive family if, for any partition of $\mathcal{X}_1,\dots,\mathcal{X}_n$ into $\mathcal{C},\bar{\mathcal{C}}$, we have
\begin{align}
\begin{split}
\forall f,x_1,\dots,x_n, \in \mathcal{V}\times\mathcal{X}_1\times\cdots\times\mathcal{X}_n, \ \ \ \\
\exists f' \in \mathcal{V} : \ \forall \bar{c}'\in\bar{\mathcal{C}}, \ f(c,\bar{c}) = f'(c,\bar{c}') \label{eqn_optional_ignorance},
\end{split}
\end{align}
where we overload $f(c,\bar{c})$ to equal $f(x_1,\dots,x_n)$ for the values of $x_1,\dots,x_n$ specified by $c,\bar{c}$.
\end{definition}
Intuitively, the constraint on $\mathcal{V}$ states that for any binary partition of the $X_1,\dots, X_n$ into known and unknown sets, if a function is expressible given some constant assignment to the unknown variables, the same function is expressible if the unknown variables are allowed to vary arbitrarily.
Intuitively, this means one can assign zero weight to those variables, so their values don't matter.
This constraint, which we refer to as \textit{multivariable optional ignorance} in reference to \citet{xu2020theory}, will be used to ensure non-negativity of information; when some $X_\ell$ is moved from $\bar{C}$ to $C$ as a new predictive variable for the agent to use, optional ignorance ensures the agent can still act as if that variable were held constant.

\begin{example}
Let $X_1,\dots, X_n \in \mathbb{R}^{d_1},\dots,\mathbb{R}^{d_n}$ and $Y\in\mathcal{Y}$ be random variables. 
Let $\Omega$ be defined as in Definition~\ref{defn_predictive_family}.
Then $V = \{f : f(x_1,\dots,x_n) = \text{softmax}(W_2\ \sigma(W_1[x_1;\cdots;x_n] + b) + b)\}$, the set of 1-layer multi-layer perceptrons, is a predictive family.
Ignorance of some $x_i$ can be achieved by setting the corresponding rows of $W_1$ to zero.
\end{example}

Given the predictive family of functions the agent has access to, we define the multivariable \vinfo analogue of entropy:
\begin{definition}[Multivariable Predictive $\mathcal{V}$-entropy]
Let $X_1,\dots, X_n\in \mathcal{X}_1,\dots,\mathcal{X}_n$.
Let $C\in\mathcal{C}$ and $\bar{C}\in\bar{\mathcal{C}}$ form a binary partition of $X_1,\dots,X_n$.
Let $\bar{a} \in \bar{\mathcal{C}}$.
Then the $\mathcal{V}$-entropy of $Y$ conditioned on $C$ is defined as
\begin{align} \label{eqn_defn_ventropy}
H_{\mathcal{V}}(Y|C) = \inf_{f \in V} \mathbb{E}_{c,y}\big[ -\log f(c,\bar{a})[y]\big].
\end{align}
\end{definition}
Note that $\bar{a}$ does not vary with $y$; thus it is `informationless'.
The notation $f(c,\bar{a})$ takes the known value of $C\subseteq \{X_1,\dots,X_n\}$, and the constant value $\bar{a}$, and produces a distribution over $\mathcal{Y}$, and $f(c,\bar{a})[y]$ evaluates the density at $y$.

If we let $V=\Omega$, the set of all functions from the $\mathcal{X}_i$ to distributions over $\mathcal{Y}$, then $\mathcal{V}$-entropy becomes exactly Shannon entropy \cite{xu2020theory}.
And just like for Shannon information, the multivariable \vinfo from some variable $X_\ell$ to $Y$ is defined as the reduction in entropy when its value becomes known.
In our notation, this means some $X_\ell$ is moving from $\bar{C}$ (the unknown variables) to $C$ (the known variables), so this definition encompasses the notion of \textit{conditional} mutual information if $C$ is non-empty to start.

\begin{definition}[Multivariable \vinfo]
Let $X_1,\dots,X_n\in\mathcal{X}_1,\dots,\mathcal{X}_n$ and $Y\in\mathcal{Y}$ be random variables. 
Let $\mathcal{V}$ be a multivariable predictive family.
Then the conditional multivariable \vinfo from $X_\ell$ to $Y$, where $\ell \in \{1,\dots,n\}$, conditioned on prior knowledge of $C \subset \{X_1,\dots,X_n\}$, is defined as
\begin{align}
I_{\mathcal{V}}(X_\ell \rightarrow Y | C) = H_{\mathcal{V}}(Y|C) - H_{\mathcal{V}}(Y|C\cup\{X_\ell\})  
\end{align}
\end{definition}

\subsection{Properties of multivariable \vinfo}
The crucial property of multivariable \vinfo as a descriptor of probing is that it can be \textit{constructed} through computation.
In the example of the agent attempting to predict the sentiment ($Y$) of an encrypted message ($X$), if the agent has $\mathcal{V}$ equal to the set of linear functions, then $I_\mathcal{V}(X\rightarrow Y)$ is small\footnote{Where $I_{\mathcal{V}}(X\rightarrow Y)$ is defined to be $I_{\mathcal{V}}(X\rightarrow Y|\{\})$ }.
A function $\phi$ that decrypts the message constructs $\mathcal{V}$-information about $Y$, since $I_{\mathcal{V}}(\phi(X)\rightarrow Y)$ is larger.
In probing, $\phi$ is interpreted to be the contextual representation learner, which is interpreted as \textit{constructing} \vinfo about linguistic properties.

\vinfo also has some desirable elementary properties, including preserving some of the properties of mutual information, like non-negativity.
(Knowing some $X_\ell$ should not reduce the agent's ability to predict $Y$).
\begin{proposition}
Let $X_1,\dots,X_n \in \mathcal{X}_1,\dots,\mathcal{X}_n$ and $Y \in \mathcal{Y}$ be random variables, and $\mathcal{V}$ and $\mathcal{U}$ be predictive families.
Let $C,\bar{C}$ be a binary partition of $X_1,\dots,X_n$.
\begin{enumerate}
  \item\textbf{Independence} If $Y, C$ are jointly independent of $X_\ell$, then $I_{\mathcal{V}}(X_\ell \rightarrow Y | C)=0$.
  \item\textbf{Monotonicity} If $\mathcal{U} \subseteq \mathcal{V}$, then $H_{\mathcal{V}}(Y | C) \leq H_{\mathcal{U}}(Y | C)$.
  \item\textbf{Non-negativity} $I_{\mathcal{V}}(X_\ell \rightarrow Y|C) \geq 0$.
\end{enumerate}
\end{proposition}

\section{Probing as Multivariable \vinfo Estimation}
We’ve described the \vinfo framework, and discussed how it captures the intuition that usable information about linguistic properties is constructed through contextualization.
In this section, we demonstrate how a small step from existing probing methodology leads to probing estimating \vinfo quantities.

\subsection{Estimating $\mathcal{V}$-entropy}
In probing, gradient descent is used to pick the function in $\mathcal{V}$ that minimizes the cross-entropy loss,
\begin{align}
\frac{1}{\dtrain} \sum_{x,y \in \dtrain} -\log p(y|\phi_i(x);\theta),
\end{align}
where $\theta$ are the trainable parameters of functions in $\mathcal{V}$.
Recalling the definition of $\mathcal{V}$-entropy, this minimization performed through gradient descent is approximating the $\inf$ over $\mathcal{V}$, since $-\log p(y|x;\theta)$ is equal to $-\log f_\theta (x)[y]$.
To summarize, this states that the supervision used in probe training can be interpreted as approximating the $\inf$ in the definition of $\mathcal{V}$-entropy.
In traditional probing, the performance of the probe is measured on the test set $\dtest$ using the traditional metric of the task, like accuracy of $F_1$ score.
In $\mathcal{V}$-information probing, we use $\dtest$ to approximate the expectation in the definition of $\mathcal{V}$-entropy. %
Thus, the performance of a single probe on representation $R$, where the performance metric is cross-entropy loss, is an estimate of $H_{\mathcal{V}}(Y|R)$.
This brings us to our framing of a probing experiment as estimating a \vinfo quantity.
\subsection{Baselined probing}
Let baselined probing be defined as in the main paper.
Then if the performance metric is defined as the negative cross-entropy loss, we have that Perf$(B)$ estimates $-H_{\mathcal{V}}(Y|B)$, Perf$(\phi(X))$ estimates $-H_{\mathcal{V}}(Y|\phi(X))$, and so baselined probing performance is an estimate of
\begin{align}
\begin{split}
&H_{\mathcal{V}}(Y|\{B\}) - H_{\mathcal{V}}(Y|\{\phi_i(X)\})\\
 &= I_{\mathcal{V}}(\phi_i(X) \rightarrow Y) - I_{\mathcal{V}}(B\rightarrow Y)
\end{split}
\end{align}
\subsection{Conditional probing}
Let conditional probing be defined as in the main paper.
Then if the performance metric is defined as the negative cross-entropy loss, we have that Perf$([B;\mathbf{0}])$ estimates $-H_{\mathcal{V}}(Y|B)$, Perf$([B;\phi(X)])$ estimates $-H_{\mathcal{V}}(Y|B, \phi(X))$, and so conditional probing performance is an estimate of
\begin{align}
\begin{split}
&H_{\mathcal{V}}(Y|\{B\}) - H_{\mathcal{V}}(Y|\{B, \phi_i(X)\})\\
&=I_{\mathcal{V}}(\phi_i(X) \rightarrow Y | B) 
\end{split}
\end{align}
The first is estimated with a probe just on $B$---under the definition of predictive family, this means providing the agent with the real values of the baseline, and some constant value like the zero vector instead of $\phi_i(X)$.
That is, holding $\bar{a} \in \phi_i(\mathcal{X})_i$ constant and sampling $b,y \sim B, Y$, the probability assigned to $y$ is $f(b,\bar{a})[y]$ for $f\in \mathcal{V}$.
The second term is estimate with a probe on both $B$ and $\phi_i(X)$. 
So, sampling $b,x,y \sim B,X,Y$, the probability assigned to $y$ is $f(b,\phi_i(x))[y]$ for $f\in\mathcal{V}$.
Intuitively, conditional probing measures the \textit{new} information in $\phi_i(X)$ because in both probes, the agent has access to $B$, so no benefit is gained from $\phi_i(X)$ supplying the same information.

\section{Proof of Proposition 1} \label{appendix_sec_proofs}

\paragraph{Monotonicity} \emph{If $\mathcal{U} \subseteq \mathcal{V}$, then $H_{\mathcal{V}}(Y | C) \leq H_{\mathcal{U}}(Y | C)$.} Proof:
\begin{equation}
    \begin{split}
        H_{\mathcal{U}}(Y|C) &= \inf_{f \in \mathcal{U}} \mathbb{E}_{c, y} \left[ - \log f[c, \bar{a}](y) \right] \\ 
        &\geq  \inf_{f \in \mathcal{V}} \mathbb{E}_{c, y} \left[ - \log f[c, \bar{a}](y) \right] \\ &= H_{\mathcal{V}}(Y|C)
    \end{split}
\end{equation}
This holds because we are taking the infimum over $\mathcal{V}$ such that if $f\in\mathcal{U}$ then $f\in\mathcal{V}$. 

\paragraph{Non-Negativity} $I_{\mathcal{V}}(X_\ell \rightarrow Y | C) \geq 0$. \emph{Where $\mathcal{V}_{\bar{C}} \subset \mathcal{V}$ is the subset of functions that satisfies $f[c, \bar{c}] = f[c, \bar{c}']\ \forall\ \bar{c}' \in \mathcal{\bar{C}}$, and $\bar{a}, \bar{a}_{/\ell}$ denote the constant values of the unknown set with and without $X_\ell$, the proof is as follows:}

\begin{equation}
    \begin{split}
        H_{\mathcal{V}}(Y|C) &= \inf_{f \in \mathcal{V}} \mathbb{E}_{c,x_\ell,y} \left[ - \log f[c,\bar{a}](y) \right] \\ 
        &= \inf_{f \in \mathcal{V}_{\bar{C}}} \mathbb{E}_{c,x_\ell,y} \left[ - \log f[c,x_\ell,\bar{a}_{/\ell}](y) \right] \\ 
        &\geq  \inf_{f \in \mathcal{V}} \mathbb{E}_{c,x_\ell,y} \left[ - \log f[c,x_\ell,\bar{a}_{/\ell}](y) \right] \\
        &= H_{\mathcal{V}}(Y|C \cup \{X_\ell\})
    \end{split}
    \label{monotonicity}
\end{equation}
By definition, $I_{\mathcal{V}}(X_\ell \rightarrow Y | C) = H_{\mathcal{V}}(Y|C) - H_{\mathcal{V}}(Y|C \cup \{X_\ell\}) \geq 0$.

\paragraph{Independence} \emph{If $Y,C$ are jointly independent of $X_\ell$, then $I_{\mathcal{V}}(X\rightarrow Y | C) = 0$. Proof:}
\begin{equation}
    \begin{split}
        H_{\mathcal{V}}&(Y|C \cup \{X_\ell\}) \\ &=  \inf_{f \in \mathcal{V}} \mathbb{E}_{c,x_\ell,y} \left[ - \log f[c,x_\ell, \bar{a}_{/\ell}](y) \right]  \\
        &= \inf_{f \in \mathcal{V}} \mathbb{E}_{x_\ell} \mathbb{E}_{c,y}  \left[ - \log f[c,x_\ell, \bar{a}_{/\ell}](y) \right] \\
        &\geq \mathbb{E}_{x_\ell} \left[ \inf_{f \in \mathcal{V}} \mathbb{E}_{c,y}  \left[ - \log f[c,x_\ell, \bar{a}_{/\ell}](y) \right] \right] \\
        &= \mathbb{E}_{x_\ell} \left[ \inf_{f \in \mathcal{V}_{\bar{C}} } \mathbb{E}_{c,y}  \left[ - \log f[c,x_\ell, \bar{a}_{/\ell}](y) \right] \right] \\
        &= \inf_{f \in \mathcal{V}_{\bar{C}} } \mathbb{E}_{c,y} \left[ - \log f[c, \bar{a}](y) \right] \\
        &\geq \inf_{f \in \mathcal{V} } \mathbb{E}_{c,y} \left[ - \log f[c, \bar{a}](y) \right] \\
        &= H_{\mathcal{V}}(Y|C)
    \end{split}
\end{equation}
In the second line, we break down the expectation based on conditional independence. Then we apply Jensen's inequality and optional ignorance to remove the expectation w.r.t. $x$. Since $\mathcal{V}_{\bar{C}} \subset \mathcal{V}$, the former's infimum is at least as large as the latter's. Then $$I_{\mathcal{V}}(X_\ell \rightarrow Y | C) = H_{\mathcal{V}}(C) - H_{\mathcal{V}}(Y|C\cup \{X_\ell\}) \leq 0$$
Combined with non-negativity (i.e., $I_{\mathcal{V}}(X_\ell \rightarrow Y | C) > 0$) we have inequality in both directions, so $I_{\mathcal{V}}(X_\ell \rightarrow Y | C) = 0$.

\section{Equivalence of \citet{xu2020theory} and our \vinfo}
In order to define conditional probing, we needed a theory of \vinfo that considered arbitrarily many predictive variables $\mathcal{X}_1, \dots, \mathcal{X}_n$.
\vinfo as presented by \citet{xu2020theory} considers only a single predictive variable $\mathcal{X}$.
It becomes extremely cumbersome, due to the use of null variables in their presentation, to expand this to more, let alone arbitrarily many variables.
So, we redefined and extended \vinfo to more naturally capture the case with an arbitrary (finite) number of variables.
In this section, we show that, in the single predictive variable case considered by \citet{xu2020theory}, our \vinfo definition is equivalent to theirs.
For the sake of this section, we'll call the \vinfo of \citet{xu2020theory} \xuvinfo, and ours \vinfo.

In particular, we show that there is a transformation from any predictive family of \xuvinfo to predictive family for \vinfo under which predictive $\mathcal{V}$-entropies are the same (and the same in the opposite direction.)

\subsection{From \citet{xu2020theory} to ours}
We recreate the definition of predictive family from \citet{xu2020theory} here:
\begin{definition}[Xu predictive family]
Let $\Upsilon = \{f : \mathcal{X} \cup \{\varnothing\} \rightarrow \mathcal{P}(\mathcal{Y})\}$.
We say that $\mathcal{U} \subseteq \Upsilon$ is a Xu predictive family if it satisfies
\begin{align}
\forall f \in \mathcal{U}, \forall P \in \text{range}(f), \exists f' \in \mathcal{U}, s.t. \\
\forall x \in \mathcal{X}, f[x]=P, f'[\varnothing] = P
\end{align}
\end{definition}

Now, we construct one of our predictive families from the Xu predictive family.
Let $\mathcal{U}\subseteq \Upsilon$ be a Xu predictive family.
We now construct a predictive family under our framework, $\mathcal{V}\subseteq \Omega$.
For each $f\in \mathcal{U}$, $f: \mathcal{X} \cup \{\varnothing\} \rightarrow \mathcal{P}(\mathcal{Y})$, construct the following two functions: first, $g$, which recreates the behavior of $f$ on the domain of $\mathcal{X}$:    
\begin{align}
&g: \mathcal{X} \rightarrow \mathcal{P}(\mathcal{Y})\\
&g : x \mapsto f(x)
\end{align}
and second, $g'$, which recreates the behavior of $f$ on $\varnothing$, given any input from $\mathcal{X}$:
\begin{align}
&g' : \mathcal{X} \rightarrow \mathcal{P}(\mathcal{Y}) \\
&g' : x \mapsto f(\varnothing)
\end{align}
Then we define our predictive family as the union of $g,g'$ for all $f\in\mathcal{V}$:
\begin{align}
\mathcal{V} = \bigcup_{f \in \mathcal{U}} \{g, g'\},
\end{align}
where $\mathcal{V}\subseteq \Omega$ and $\Omega = \{ f: \mathcal{X} \rightarrow \mathcal{P}(\mathcal{Y})\}$.
Note from this construction that we've eliminated the presence of the null variable from the definition of predictive family.

We now show that $\mathcal{V}$, as defined in the construction above, is in fact a predictive family under our definition.
Under our definition, there are two cases: either $\mathcal{X}\in C$ or $\mathcal{X} \in \bar{C}$.
If $\mathcal{X} \in C$, then for all $f, x \in \mathcal{V} \times \mathcal{X}$, if we take any $f' \in \mathcal{V}$ (which is non-empty), then there is no $\bar{c}'\in\bar{C}$, so vacuously the condition holds.
If $\mathcal{X} \in \bar{C}$, then for all $f, x \in \mathcal{V}$, we have that $f$ was either $g$ or $g'$ for some function $h\in\mathcal{U}$ in the construction of $\mathcal{V}$ (because all functions in $\mathcal{V}$ were part of some pair $g$, $g'$.)
Then we take $f'=g'$, and have that for all $\bar{c}' \in \bar{C}$, that is $x\in\mathcal{X}$, $f'(c,\bar{c}) = f'(c,\bar{c}') = g'(x) = h(\varnothing)$, satisfying the constraint.

Finally, we show that the predictive $\mathcal{V}$-entropies of $\mathcal{V}$ (under our definition) and $\mathcal{U}$ (under that of \citet{xu2020theory}) are the same.
Consider Xu-predictive entropies: 
\begin{align}
&H_{\mathcal{U}}(Y|X) = \inf_{f\in\mathcal{U}} \mathbb{E}_{x,y} [-\log f[x](y)] \label{eqn_xu_ventropy1}\\
&H_{\mathcal{U}}(Y|\varnothing) = \inf_{f\in\mathcal{U}} \mathbb{E}_{y} [-\log f[\varnothing](y)] \label{eqn_xu_ventropy2}
\end{align}

First we want to show  $H_{\mathcal{U}}(Y|X) = H_{\mathcal{V}}(Y|X)$.
Consider the $\inf$ in Equation~\ref{eqn_xu_ventropy1}; the $f\in\mathcal{U}$ that achieves the $\inf$ corresponds to some $g\in\mathcal{V}$ by construction, and since $f(x) = g(x)$, we have that the value of the inf for $\mathcal{V}$ is at least as low as for $\mathcal{U}$.
The same is true in the other direction; in our definition $H_{\mathcal{V}}(Y|\mathcal{X})$, the $g$ that achieves the $\inf$ corresponds to some $f\in\mathcal{U}$ that produces the same probability distributions.
So, $H_{\mathcal{U}}(Y|X) = H_{\mathcal{V}}(Y|X)$.

Now we want to show  $H_{\mathcal{U}}(Y|\varnothing) = H_{\mathcal{V}}(Y)$.
Now, consider the $\inf$ in Equation~\ref{eqn_xu_ventropy2}.
The $f\in\mathcal{U}$ that achieves the $\inf$ corresponds to some $g,g'$ in the construction of $\mathcal{V}$; that $g'$ takes any $x\in\mathcal{X}$ and produces $f[\varnothing]$; hence the value of the inf for $\mathcal{V}$ is at least as low as for $\mathcal{U}$.
The same is true in the other direction. We have that $H_{\mathcal{V}}(Y) = \inf_{f\in\mathcal{V}}\mathbb{E}_{y}[-\log f(\bar{a})[y]]$ for any $\bar{a}\in\mathcal{X}$.
Either a $g$ or a $g'$ from the construction of $\mathcal{V}$ achieves this $\inf$; if a $g$ achieves it, then its corresponding $g'$ emits the same probability distributions, so WLOG we'll assume it's a $g'$.
We know that $g'(\bar{a})=f(\varnothing)$ for all $\bar{a}\in\mathcal{X}$, so $H_{\mathcal{U}}(Y|\varnothing)$ is at most $H_{\mathcal{V}}(Y)$.
So, $H_{\mathcal{U}}(Y|\varnothing) = H_{\mathcal{V}}(Y)$.

Since the V-entropies of the predictive family from \citet{xu2020theory} and ours are the same, all the information quantities are the same.
This shows that the predictive family we constructed in our theory is equivalent to the predictive family from \citet{xu2020theory} that we started with.

\subsection{From our \vinfo to that of \citet{xu2020theory}}
Now we construct a predictive family $\mathcal{U}$ under the framework of \citet{xu2020theory} from an arbitrary predictive family $\mathcal{V}$ under our framework.
For each function $f\in\mathcal{V}$, we have from the definition that there exists $f'\in\mathcal{V}$ such that $\forall x\in\mathcal{X}$, $f'(x) = P$ for some $P\in\mathcal{P}(\mathcal{Y})$.
We then define the function:
\begin{align}
 g : \mathcal{X} \cup \{\varnothing\} \rightarrow \mathcal{P}(\mathcal{Y})\\
 g(x) = \begin{cases}
 f(x) & x \in \mathcal{X} \\
 f'(\bar{a}) & x =\varnothing 
 \end{cases}
\end{align}
where $\bar{a}\in\mathcal{X}$ is an arbitrary element of $\mathcal{X}$, and the set of constant-valued functions
\begin{align}
G = \{ g': g'(z) = P \mid P \in \text{range}(f)\},
\end{align}
where $z\in\mathcal{X} \cup \{\varnothing\}$, and let 
\begin{align}
    \mathcal{U} = \bigcup_{f\in\mathcal{V}} \{g\}\cup G
\end{align}
The set $\mathcal{U}$ is a predictive family under Xu-\vinfo because for any $f\in\mathcal{U}$, $f$ is either a $g$ or a $g'$ in our construction, and so optional ignorance is maintained by the set $G$ that was either constructed for $g$ or that $g'$ was a part of.
That is, from the construction, $G$ contains a function for each element in the range of $g$ (or $g'$) that maps all $x\in\mathcal{X}$ as well as $\varnothing$ to that element, and $\mathcal{U}$ contains all elements in $G$.

Now we show that the predictive $\mathcal{V}$-entropies of $\mathcal{U}$ (from this construction) under \citet{xu2020theory} are the same as for $\mathcal{V}$ under our framework.

First we want to show  $H_{\mathcal{U}}(Y|X) = H_{\mathcal{V}}(Y|X)$.
For the $g$ that achieves the $\inf$ over $\mathcal{U}$ in Equation~\ref{eqn_xu_ventropy1}, we have there exists $f\in\mathcal{V}$ such that $g(x) = f(x)$ given that $x\in\mathcal{X}$, so $H_{\mathcal{V}}(y|x) \leq H_{\mathcal{U}}(y|x)$
The same is true in the other direction; the $f\in\mathcal{V}$ that achieves the $\inf$ in $\mathcal{V}$-entropy similarly corresponds to $g\in\mathcal{U}$, implying $H_{\mathcal{U}}(Y|X) \leq H_{\mathcal{V}}(Y|X)$, and thus their equality.

Now we want to show  $H_{\mathcal{U}}(Y|\varnothing) = H_{\mathcal{V}}(Y)$.
For the $g\in\mathcal{U}$ that achieves its $\inf$, we have by construction that there is an $f' \in \mathcal{V}$ such that for any $\bar{a} \in \mathcal{X}$, it holds that $g(\varnothing) = f'(\bar{a})$.
So, $H_{\mathcal{V}}(Y|X) \leq H_{\mathcal{U}}(Y|X)$.
In the other direction, for the $f\in\mathcal{V}$ that achieves its $\inf$ given an arbitrary $\bar{a}\in \mathcal{X}$, there is the $f'\in\mathcal{V}$ from our construction of $\mathcal{U}$ such that $f(\bar{a}) = f'(x) = g(\varnothing)$ for all $x\in\mathcal{X}$.
This implies $H_{\mathcal{U}}(Y|X) \leq H_{\mathcal{V}}(Y|X)$, and thus their equality. 

\subsection{Remarks on the relationship between our \vinfo and that of \citet{xu2020theory}} \label{appendix_section_xu_ours_relationship}

The difference between our \vinfo and that of \citet{xu2020theory} is in how the requirement of \textit{optional ignorance} is encoded into the formalism.
This is an important yet technical requirement that if a predictive agent has access to the value of a random variable $X$, it's allowed to \textit{disregard} that value if doing so would lead to a lower entropy.
An example of a subset of $\Omega$ for which this \textit{doesn't} hold in the multivariable case is for multi-layer perceptrons with a frozen (and say, randomly sampled) first linear transformation.
The information of, say, $X_1$ and $X_2$, are mixed by this frozen linear transformation, and so $X_1$ cannot be ignored in favor of just looking at $X_2$.
However, if the first linear transformation is trainable, then it can simply assign $0$ weights to the rows corresponding to $X_1$ and thus ignore it.

The \vinfo of \citet{xu2020theory}  ensures this option by introducing a null variable $\varnothing$ which is used to represent the lack of knowledge about their variable $X$ -- and for any probability distribution in the range of some $f\in\mathcal{U}$ under the theory, there must be some function $f$ that produces the same probability distribution when given any value of $X$ or $\varnothing$.
This is somewhat unsatisfying because $f$ should really be a function from $\mathcal{X} \rightarrow \mathcal{P}(\mathcal{Y})$, but this implementation of optional ignorance changes the domain to $\mathcal{X} \cup \{\varnothing\}$. 
When attempting to extend this to the multivariable case, the definition of optional ignorance becomes very cumbersome.
With two variables, the domain of functions in a predictive family must be $(\mathcal{X}_1 \cup \{\varnothing\}) \times (\mathcal{X}_2 \cup \{\varnothing\})$.
Because the definition of $\mathcal{V}$-entropy under \citet{xu2020theory} treats using $\mathcal{X}$ separately from using $\varnothing$, one must define optional ignorance constraints separately for each subset of variables to be ignored, the number of which grows exponentially with the number of variables.

Our re-definition of \vinfo gets around this issue by defining the optional ignorance constraint in a novel way, eschewing the $\varnothing$ and instead encoding it as the intuitive implementation that we described in the MLP -- that for any function in the family and \textit{fixed} value for some subset of the inputs (which will be the unknown subset), there's a function that behaves identically even if that subset of values is allowed to take \textit{any} value.
(Intuitively, by, e.g., having it be possible that the weights for those inputs are $0$ at the first layer.)

\section{Full Results}

\begin{figure}
\includegraphics[width=\linewidth]{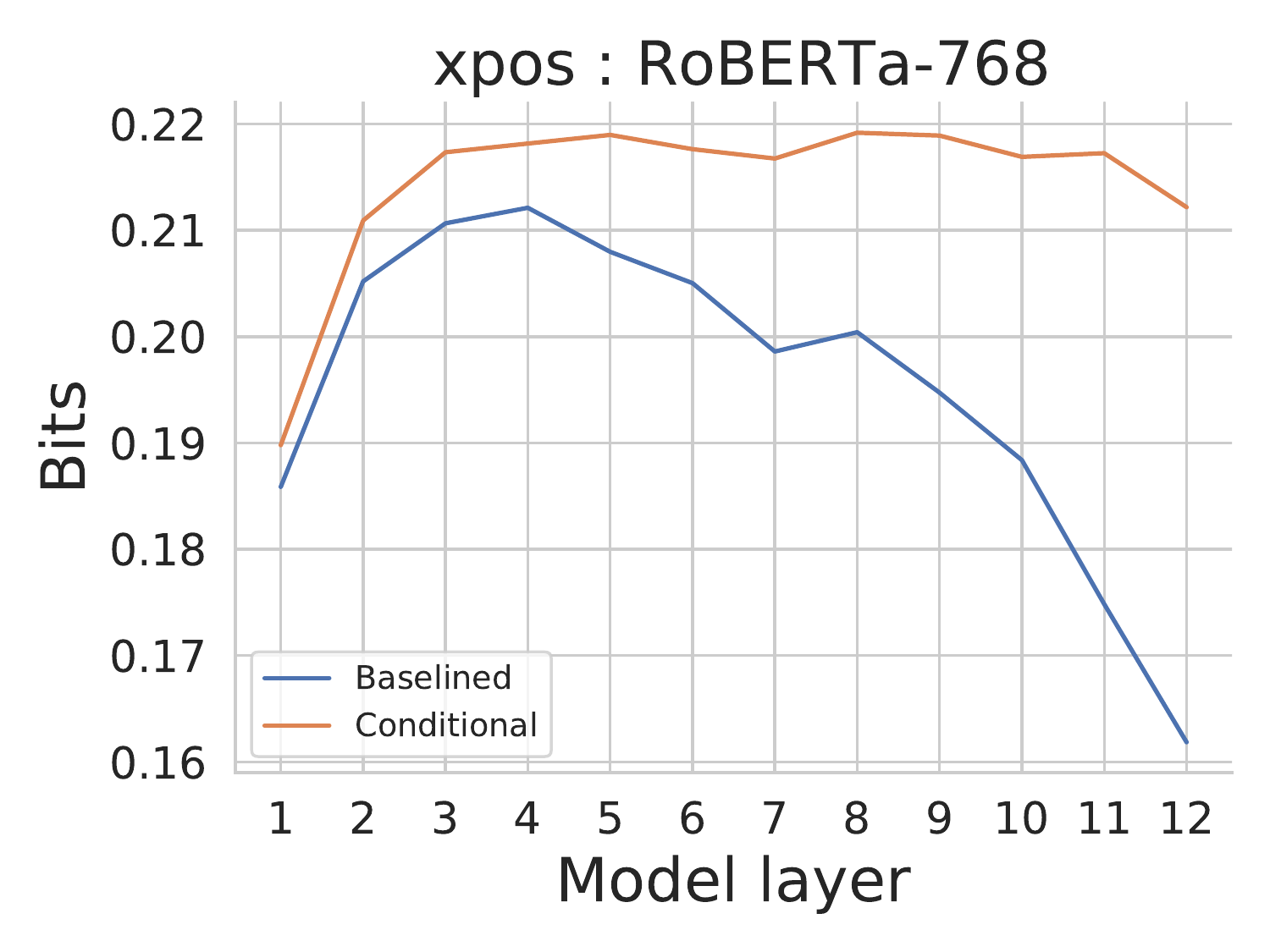}
\caption{\label{appendix_figure_roberta_xpos}Probing results on RoBERTa for xpos. Results are reported in bits of $\mathcal{V}$-information; higher is better}
\end{figure}

In this section, we report all individual probing experiments: single-layer probes' $\mathcal{V}$-entropies in Table~\ref{appendix_results_single_ventropy}, single-layer probes' task-specific metrics in Table~\ref{appendix_results_single_metrics}, two-layer probes' $\mathcal{V}$-entropies in Table~\ref{appendix_results_two_ventropy}, and two-layer probes' task-specific metrics in Table~\ref{appendix_results_two_metrics}.
In Figure~\ref{appendix_figure_roberta_xpos}, we report the xpos figure for RoBERTa corresponding to the other four figures in the main paper.
We see that it shows roughly the same trend as the upos figure from the main paper.

\section*{Acknowledgements}
The authors would like to thank the anonymous reviewers and area chair for their helpful feedback which led to a stronger final draft, as well as Nelson Liu, Steven Cao, Abhilasha Ravichander, Rishi Bommasani, Ben Newman, and Alex Tamkin, and much of the Stanford NLP Group for helpful discussions and comments on earlier drafts.
JH was supported by an NSF Graduate Research Fellowship under grant number DGE-1656518, and by Two Sigma under their 2020 PhD Fellowship Program. KE was supported by an NSERC PGS-D scholarship.

\begin{table}
  \centering
  \small
  \begin{tabular}{c c c c c c}
&\multicolumn{5}{c}{RoBERTa Single-Layer $\mathcal{V}$-Entropy}\\
  \toprule
Layer & upos & xpos & dep & ner & sst2\\
\midrule
0&0.336 &0.344 &1.468 &0.391 &0.643\\
1&0.145 &0.158 &0.827 &0.216 &0.645\\
2&0.119 &0.139 &0.676 &0.188 &0.630\\
3&0.118 &0.133 &0.635 &0.172 &0.574\\
4&0.117 &0.132 &0.627 &0.167 &0.545\\
5&0.119 &0.136 &0.632 &0.167 &0.489\\
6&0.121 &0.139 &0.645 &0.167 &0.484\\
7&0.126 &0.145 &0.640 &0.170 &0.462\\
8&0.129 &0.144 &0.633 &0.168 &0.467\\
9&0.131 &0.149 &0.653 &0.173 &0.494\\
10&0.138 &0.156 &0.677 &0.177 &0.508\\
11&0.154 &0.169 &0.705 &0.184 &0.527\\
12&0.161 &0.182 &0.746 &0.191 &0.583\\
\bottomrule
\end{tabular}
\caption{\label{appendix_results_single_ventropy}
$\mathcal{V}$-entropy results (in bits) on probes taking in one layer, for each layer of the network. Lower is better.}
\end{table}

\begin{table}
  \centering
  \small
  \begin{tabular}{c c c c c c}
&\multicolumn{5}{c}{RoBERTa Single-Layer Metrics}\\
  \toprule
Layer & upos & xpos & dep & ner & sst2\\
\midrule
0  &0.908 &0.908 &0.669 &0.535 &0.808\\
1  &0.968 &0.964 &0.821 &0.710 &0.815\\
2  &0.975 &0.969 &0.854 &0.735 &0.813\\
3  &0.975 &0.970 &0.865 &0.763 &0.845\\
4  &0.976 &0.971 &0.867 &0.763 &0.850\\
5  &0.975 &0.970 &0.866 &0.763 &0.869\\
6  &0.975 &0.970 &0.863 &0.764 &0.870\\
7  &0.974 &0.969 &0.864 &0.754 &0.877\\
8  &0.974 &0.970 &0.865 &0.762 &0.868\\
9  &0.974 &0.969 &0.863 &0.756 &0.860\\
10 &0.973 &0.968 &0.859 &0.756 &0.857\\
11 &0.971 &0.967 &0.854 &0.744 &0.850\\
12 &0.969 &0.965 &0.847 &0.735 &0.843\\
\bottomrule
\end{tabular}
\caption{\label{appendix_results_single_metrics}
Task-specific metric results on probes taking in one layer, for each layer of the network. For upos, xpos, dep, and sst2, the metric is accuracy. For NER, it's span-level $F_1$ as computed by the Stanza library \cite{qi2020stanza}. For all metrics, higher is better.}
\end{table}

\begin{table}
  \centering
  \small
  \begin{tabular}{c c c c c c}
&\multicolumn{5}{c}{RoBERTa Two-Layer $\mathcal{V}$-entropy}\\
  \toprule
Layer & upos & xpos & dep & ner & sst2\\
\midrule
0-0  &0.335 &0.345 &1.466 &0.391 &0.639\\
0-1  &0.141 &0.154 &0.763 &0.210 &0.633\\
0-2  &0.115 &0.133 &0.646 &0.184 &0.615\\
0-3  &0.110 &0.127 &0.609 &0.169 &0.567\\
0-4  &0.109 &0.126 &0.593 &0.164 &0.549\\
0-5  &0.110 &0.125 &0.602 &0.163 &0.474\\
0-6  &0.109 &0.126 &0.613 &0.165 &0.484\\
0-7  &0.109 &0.127 &0.609 &0.166 &0.451\\
0-8  &0.108 &0.125 &0.598 &0.171 &0.462\\
0-9  &0.109 &0.125 &0.614 &0.167 &0.489\\
0-10 &0.110 &0.127 &0.636 &0.177 &0.504\\
0-11 &0.111 &0.127 &0.654 &0.175 &0.525\\
0-12 &0.116 &0.132 &0.682 &0.185 &0.563\\
\bottomrule
\end{tabular}
\caption{\label{appendix_results_two_ventropy}
$\mathcal{V}$-entropy results on probes taking in two layers: layer $0$ and each other layer of the network. Lower is better.}
\end{table}

\begin{table}
  \centering
  \small
  \begin{tabular}{c c c c c c}
&\multicolumn{5}{c}{RoBERTa Two-Layer Metrics}\\
  \toprule
Layer & upos & xpos & dep & ner & sst2\\
\midrule
0-0  &0.908 &0.907 &0.670 &0.543 &0.808\\
0-1  &0.969 &0.965 &0.834 &0.722 &0.825\\
0-2  &0.975 &0.970 &0.861 &0.744 &0.822\\
0-3  &0.976 &0.971 &0.870 &0.765 &0.850\\
0-4  &0.977 &0.972 &0.874 &0.767 &0.849\\
0-5  &0.977 &0.972 &0.872 &0.772 &0.875\\
0-6  &0.977 &0.972 &0.869 &0.766 &0.875\\
0-7  &0.977 &0.972 &0.869 &0.760 &0.869\\
0-8  &0.977 &0.972 &0.872 &0.762 &0.862\\
0-9  &0.977 &0.972 &0.869 &0.766 &0.864\\
0-10 &0.977 &0.972 &0.864 &0.755 &0.857\\
0-11 &0.977 &0.972 &0.861 &0.753 &0.859\\
0-12 &0.975 &0.971 &0.856 &0.743 &0.847\\
\bottomrule
\end{tabular}
\caption{\label{appendix_results_two_metrics}Task-specific metric results on probes taking in two layers: layer $0$ and each other layer of the network. For upos, xpos, dep, and sst2, the metric is accuracy. For NER, it's span-level $F_1$ as computed by the Stanza library. For all metrics, higher is better.}
\end{table}

\end{document}